\documentclass[11pt]{article}
\usepackage{ranlp2015}
\usepackage{multirow}
\usepackage{times}
\usepackage{url}
\usepackage{graphicx,epstopdf}
\usepackage{latexsym}
\usepackage[tight]{subfigure}
\usepackage{algorithm}
\usepackage{algorithmic}
\usepackage{amsmath} 
\usepackage{amssymb}  
\usepackage{cases}
\usepackage{float}
\usepackage{array}
\newcolumntype{C}[1]{>{\centering}p{#1}}

\DeclareMathOperator*{\argmin}{arg~min}

\title{Jointly Embedding Relations and Mentions for Knowledge Population}

\author{Miao Fan$^{\dagger,\ddagger,*}$, Kai Cao$^{\ddagger}$, Yifan He$^{\ddagger}$ and Ralph Grishman$^{\ddagger}$\\
$^{\dagger}$ CSLT, Division of Technical Innovation and Development,\\ Tsinghua National Laboratory for Information Science and Technology,\\ Tsinghua University, Beijing, 100084, China.\\
$^{\ddagger}$Proteus Group, New York University, NY, 10003, U.S.A.
}
\date{}

\begin{document}
\maketitle
\begin{abstract}
This paper contributes a joint embedding model for predicting relations between a pair of entities in the scenario of relation inference. It differs from most stand-alone approaches which separately operate on either knowledge bases or free texts. The proposed model simultaneously learns low-dimensional vector representations for both triplets in knowledge repositories and the mentions of relations in free texts, so that we can leverage the evidence both resources to make more accurate predictions. We use NELL to evaluate the performance of our approach, compared with cutting-edge methods. Results of extensive experiments show that our model achieves significant improvement on relation extraction.
\end{abstract}

\section{Introduction}
Relation extraction \cite{bach2007review,Grishman:1997:IET:645856.669801,sarawagi2008information}, which aims at discovering the relationships between a pair of entities, is a significant research direction for discovering more beliefs for knowledge bases. Most stand-alone approaches, however, either use local graph patterns in knowledge repositories, or extract features from text mentions, to individually help predict relations between two entities. The heterogeneity brings about a gap between structured repositories and unstructured free texts, which spoils the dream of sharing the evidence from both knowledge and natural language.

For studies in decades, scientists either compete the performance of their methods on the public text datasets such as ACE\footnote{http://www.itl.nist.gov/iad/mig/tests/ace/} \cite{GuoDong:2005:EVK:1219840.1219893} and MUC\footnote{http://www.itl.nist.gov/iaui/894.02/related projects/muc/} \cite{zelenko2003kernel}, or look for effective approaches \cite{conf/emnlp/GardnerTKM13,lao-mitchell-cohen:2011:EMNLP} on improving the accuracy of link prediction within knowledge bases such as NELL\footnote{http://rtw.ml.cmu.edu/rtw/} \cite{carlson-aaai} and Freebase\footnote{http://www.freebase.com/} \cite{Bollacker2007}.
Thanks to the research of distantly supervised relation extraction  \cite{fan-EtAl:2014:P14-1,mintz2009distant} which facilitates the manual annotation via automatically aligning with the relation mentions in free texts, NELL can not only extract triplets, i.e. $\langle head\_entity, relation, tail\_entity\rangle$, but also collect the texts between two entities as the evidence of relation mention. We take an example from NELL which originally records a belief:
$\langle concept:city:caroline,	concept:citylocatedinstate,	concept:stateorprovince:maryland,	 County~and~State~of\rangle$, where  ``$County~and~State~of$'' is the mention between the head entity $concept:city:caroline$, and the tail entity $concept:stateorprovince:maryland$, to indicate the relation $concept:citylocatedinstate$.

Fortunately, the embedding techniques \cite{Fan-EtAl:2014:PACLIC,Mikolov2013} enlighten us to break through the limitation of heterogeneous resources, and to establish a connection between a relation and its corresponding mention via learning a specific vector representation for each of the elements, including the entities and relations in triplets, and the words in mentions. More specifically, we propose a joint relation mention embedding (JRME) model in this paper, which simultaneously learns low-dimensional vector representations for entities and relations in knowledge repositories, and in the meanwhile, each word in the relation mentions is also trained a dedicated embedding. This model helps us take advantage of the benefits from the two resources to make more accurate predictions. We use two different datasets extracted from NELL to evaluate the performance of JRME, compared with cutting-edge methods. It turns out that our model achieves significant improvement on relation extraction.
\section{Related Work}
\begin{figure*}
  \centering
  \includegraphics[width=0.95\textwidth]{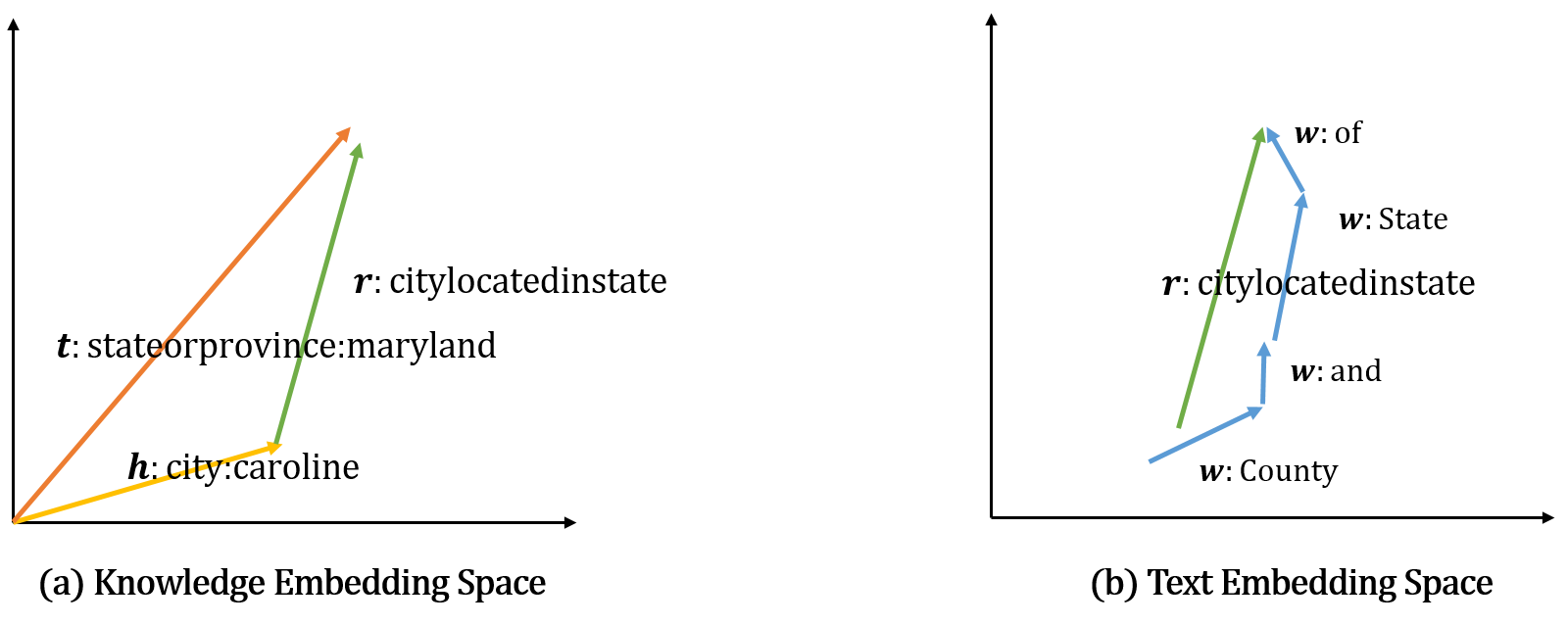}\\

  \caption{Given a belief, ${\bf h}: city:caroline,	{\bf r}:citylocatedinstate,	{\bf t}:stateorprovince:maryland$ and ${\bf m}:	 County~and~State~of$ in NELL, (a) shows the distributed representations of a triplet in the knowledge space, and (b) illustrates word embeddings in the text space.}
\end{figure*}

We group some recent work on relation extraction into two categories, i.e. text-based approaches and knowledge-based methods. Generally speaking, both of the parties seek better evidences to make more accurate predictions. The text-based community focuses on linguistic features such as the words combined with POS tags that indicate the relations, but the other side conducts relation inference depending on the local connecting patterns between entity pairs learnt from the knowledge graph which is established by beliefs.
\subsection{Text-based Approaches}
It is believed that the text between two recognized entities in a sentence indicate their relationships to some extent. To implement a relation extraction system guided by supervised learning, a key step is to annotate the training data. Therefore, two branches emerge as follows,

\begin{itemize}
  \item {\it Relation extraction with manual annotated corpora}: Traditional approaches compete the performance on the public text datasets which are annotated by experts, such as ACE and MUC. They choose different features extracted from the texts, like kernel features \cite{zelenko2003kernel} or semantic parser features \cite{GuoDong:2005:EVK:1219840.1219893},
      and there is a comprehensive survey \cite{sarawagi2008information} which shows more details about this branch.
  \item {\it Relation extraction with distant supervision}: Due to the limited scale and tedious labor caused by manual annotation, scientists explore an alternative way to automatically generate large-scale annotated corpora, named by distant supervision \cite{mintz2009distant}. Even though this cutting-edge technique solves the issue of lacking annotated corpora, we still suffer from the problem of noisy and sparse features \cite{fan-EtAl:2014:P14-1}.
\end{itemize}

\subsection{Knowledge-based Methods}
Knowledge bases contain millions of entries which are usually represented as triplets, i.e. $\langle head\_entity, relation, tail\_entity\rangle$, which intuitively inspire us to regard the whole repository as a graph, where entities are nodes and relations are edges. Therefore, one research community looks forward to predicting unknown relations which may exist between two entities via learning the linking patterns, and another promising research group tries to learn structured embeddings of knowledge bases.

\begin{itemize}
  \item {\it Relation prediction with graph patterns}: Some canonical studies  \cite{conf/emnlp/GardnerTKM13,lao-mitchell-cohen:2011:EMNLP} adopt a data-driven random walk model, which follows the paths from the head entity to the tail entity on the local graph structure to generate non-linear feature combinations to represent relations, and then uses logistic regression to select the significant features that contribute to classifying other entity pairs which also have the given relation.

  \item {\it Relation prediction with embedding representations}: Bordes et al. \cite{Bordes2013a, Bordes2011} propose an alternative way that embedding the whole knowledge graph via learning a specific low-dimensional vector for each entity and relation, so that we just need simple vector calculation instead to predict relations.
\end{itemize}

Our model (JRME) benefits more from the latest and state-of-the art embedding approaches, TransE \cite{Bordes2013a} and IIKE \cite{fan2015learning}. Therefore, we re-implement them as the rival methods, and conduct extensive comparisons in the subsequent experiments.

\section{Model}
The heterogeneity between free texts and knowledge bases brings about a challenge that we can hardly take advantage of the features uniformly, since they are located in different spaces and have varies dimensions.
Thankfully, the embedding techniques \cite{Fan-EtAl:2014:PACLIC,Mikolov2013,fan2015probabilistic,fanprobabilistic} leave an idea that almost all the elements, including words, entities, relations, can be learnt and assigned distributed representations, and the mission remaind for us is to jointly learn embeddings for entities, relations, and the words in the same feature space.

We arrange the subsequent content as follows: Section 3.1 and 3.2 describe how to model the knowledge and texts individually, and we finally talk about the proposed jointly embedding model in Section 3.3.

\subsection{Knowledge Relation Embedding}
Inspired by TransE \cite{Bordes2013a}, we regard the relation $r$ between a pair of entities, i.e. $h$ and $t$, as a transition, due to the hierarchical structure of knowledge graphs. Therefore, we use $D_r(h,r,t)$ as follows to denote the plausibility of a triplet $(h,r,t)$ illustrated by Figure 1(a):
\begin{equation}
D_r(h,r,t) = |{\bf h} + {\bf r} - {\bf t}|^2,
\end{equation}
where the closer ${\bf h} + {\bf r}$ is to ${\bf t}$, the more likely the triplet $(h,r,t)$ exists. The bold fonts indicate the vector representations, e.g. the embedding of the head entity $h$ is ${\bf h} \in \mathbb{R}^d$ where $d$ is short for dimension.

Assume that $R$ is the set of relations. Given a correct triplet $(h,r,t)$, we aim at pushing all the possible corrupt triplets with wrong relations $\{r'| r' \in R ~{\&} ~r' \not= r\}$ away. Therefore, we adopt a margin-based ranking loss function with a block $\alpha$ to separate all the negative triplets in the corrupted base $K'$ from all the positives in the correct knowledge base $K$:
\begin{equation}
\begin{split}
& \argmin_{r, r'} ~\mathcal{L}_r = \sum_{(h, r, t) \in K} \sum_{(h, r', t) \in K'} [\alpha + D_r(h, r, t)\\
&- D_r(h, r', t)]_+,\\
\end{split}
\end{equation}
in which $[~]_+$ is a hinge loss function, i.e. $[x]_+ = {\text max}(0, x)$.
\subsection{Text Mention Embedding}
Similar to the Knowledge Relation Embedding (KBE), we can also find an approach to measure the distance between the mention $m$ and its corresponding relation $r$ in Text Mention Embedding (TME). To denote the embedding of mention ${\bf m}$, we sum all the embeddings of words included by $m$ as shown by Equation (3). Thanks to representing all the words and relations in vectors with the same dimension which is demonstrated by Figure 1(b), we can adopt inner product function shown by Equation (4) to calculate their similarity.
\begin{equation}
{\bf m} = \sum_{w \in m} {\bf w},
\end{equation}
\begin{equation}
D_m(r, m) = - {\bf r}^T {\bf m}.
\end{equation}
Before using the margin-based ranking loss function to learn, we need to construct the negative set $T'$ for each pair of relation mention $(r, m)$ which appears in the correct training set $T$. To generate the negative pairs $(r', m)$, we keep the mention $m$ but iteratively change other relations from the set of relations $R$. The subsequent Formula (5) helps to discriminate between the two opponent sets with a margin $\beta$,
\begin{equation}
\begin{split}
&\argmin_{r, m, r'} ~\mathcal{L}_m = \sum_{(r, m) \in T} \sum_{(r', m) \in T'} [\beta + D_m(r, m) \\
& - D_m(r', m)]_+.\\
\end{split}
\end{equation}
\subsection{Joint Relation Mention Embedding}
Due to the uniform modeling standard of KBE and TME, we can jointly embed the relations and corresponding mentions (JRME) with Equation (6),
\begin{equation}
\begin{split}
&\argmin_{r, m, r'} ~\mathcal{L} = \sum_{(h, r, t, m) \in KT} \sum_{(h, r',t, m) \in KT'}  [\gamma \\
&+ D_r(h, r, t) - D_r(h, r', t)\\
&+D_m(r, m) - D_m(r', m)]_+,\\
\end{split}
\end{equation}
in which each belief $(h, r, t, m)$ belonging to the training set $KT$ contains two entities, the relation and its corresponding mention.

If we achieve the learnt embeddings for all the entities, relations and words in mentions, we can simply use Equation (7) to measure the rationality of a relation $r$ appearing between a pair of entities $h, t$ with the evidence of $m$:
\begin{equation}
Score(h, r, t, m) = D_r(h, r, t) + D_m(r, m)
\end{equation}

\section{Experiments}
We set up three objectives for evaluating the effectiveness of JRME, which are:

\begin{itemize}
  \item testing the effectiveness of JRME in terms of different evaluation protocols/metrics;
  \item comparing the performances of JRME with other cutting-edge approaches;
  \item judging the robustness of the proposed model by using a larger but noisy dataset.
\end{itemize}

Section 4.1 and 4.2 display the different datasets and the various protocols we use to measure the performance compared with several state-of-the-art approaches, i.e TransE \cite{Bordes2013a} and IIKE \cite{fan2015learning}. Section 4.3 will show the results of the extensive experiments.

\subsection{Datasets}
We prepare two datasets with different statistical characteristics. As illustrated by Table 1, both of them are generated by NELL \cite{carlson-aaai}, a Never-Ending Language Learner which works on automatically extracting beliefs from the Web. NELL-50K is a medium size dataset, and each belief, which contains the head entity $h$, the tail entity $t$, the relation $r$ between them, and the mention $m$ indicate the relation, is validated by experts. However, NELL-5M is a much larger one with five million uncertain training examples automatically learnt from the Web by NELL.
\begin{table}
\centering
\begin{tabular}{|c|c|c|}
  \hline
  {\bf DATASET} & {\bf NELL-50K} & {\bf NELL-5M}\\
  \hline
  \hline
   \#(ENTITIES) & 29,904  & 177,635 \\
  \#(RELATIONS) & 233  & 236 \\
  \#(TRAINING EX.) & 57,356  & 5,000,000\\
  \#(VALIDATING EX.) & 10,710 & 47,335 \\
  \#(TESTING EX.) & 10,711 & 47,335 \\
  \hline
\end{tabular}
\caption{Statistics of the datasets used for relation prediction task.}
\end{table}

\subsection{Protocols}
The scenario of experiments is that: given a pair of entities, a short text/mention to indicate the correct relations and a set of candidate relations, we compare the performance between our models and other state-of-the-art approaches, with the metrics as follows,

\begin{itemize}
  \item {\it Average Rank}: Each candidate relation will gain a score calculated by Equation (7). We sort them in ascending order and compare with the corresponding ground-truth belief. For each belief in the testing set, we get the rank of the correct relation. The average rank is an aggregative indicator, to some extent, to judge the overall performance on relation extraction of an approach.
  \item {\it Hit@10}: Besides the average rank, scientists from the industrials concern more about the accuracy of extraction when selecting Top10 relations. This metric shows the proportion of beliefs that we predict the correct relation ranked in Top10.
  \item {\it Hit@1}: It is a more strict metric that can be referred by automatic system, since it demonstrates the accuracy when just picking the first predicted relation in the sorted list.
\end{itemize}

\subsection{Hyperparameters}
Before displaying the evaluation results, we need to elaborate the hyperparameters that have been tried, and show the best combination of hyperparameters we choose.
Another advantage of embedding-based model is that it is unnecessary to tune many hyperparameters. For our model, we just need to set four, which are the uniform dimension $d$ of entities, relations and the words in mentions, the margin $\alpha$ of KBE, the margin $\beta$ of TME and the margin $\gamma$ of JRME. To decide the ideal set of hyperparameters, we use the validation set to pick the best combination from
$d \in \{10, 20, 50, 100, 200\}$,
$\alpha \in \{0.1, 1.0, 2.0, 5.0, 10.0\}$,
$\beta \in \{0.1, 1.0, 2.0, 5.0, 10.0\}$ and
$\gamma \in \{0.1, 1.0, 2.0, 5.0, 10.0\}$. Finally, we choose $d = 100, \alpha = 1.0, \beta = 1.0$ and $\gamma = 2.0$ to train the embeddings, as this combination of hyperparameters helps perform best on the validation set.
\subsection{Performance}
Table 2 and 3 illustrate the results of experiments on NELL-50K and NELL-5M, respectively. Both of them show that JRME performs best among all the approaches we implemented. We can also figure out that text mentions contribute a lot to predicting the correct relations.
Moreover, Table 3 also demonstrates that not only IIKE is robust to the noise in NELL-5M dataset, which consists with its characteristics emphasized by Fan et al. \cite{fan2015learning}, but also TME and JRME share this special ``gene''. Overall, JRME improves the average rank of relation prediction about 20\% compared with state-of-the-art IIKE.

\begin{table}
\centering
\begin{tabular}{|c|c|c|c|c|}
  \hline
  {\bf APPROACH} & {\bf AVG. R.} & {\bf HIT@10} & {\bf HIT@1}  \\
  \hline
  \hline
   TransE   & 131.8 & 16.3\% & 3.0\% \\
   KRE & 29.1 & 44.3\% & 14.4\%\\
   TME & 11.5 & 80.0\% & 56.0\%  \\
  IIKE  & {\it 7.5} & {\it 81.8\%} & {\it 56.8\%}\\

  JRME & {\bf 6.2} & {\bf 87.8\%}  & {\bf 60.2\%} \\
  \hline
\end{tabular}
\caption{Performance of TransE, KRE, IIKE, TME and JRME on the metrics of Average Rank, Hit@10 and Hit@1 in NELL-50K dataset.}
\end{table}

\begin{table}
\centering
\begin{tabular}{|c|c|c|c|c|}
  \hline
  {\bf APPROACH} & {\bf AVG. R.} & {\bf HIT@10} & {\bf HIT@1}  \\
  \hline
  \hline
   TransE   & 77.1 & 5.4\% & 0.7\% \\
   KRE & 57.5 & 17.9\% & 2.5\%\\
   TME & {\it 3.6} & {\it 96.3\%} & {\it 63.6\%}  \\
  IIKE   & 4.5 & 82.6\% & 53.2\%\\

  JRME & {\bf 3.0} & {\bf 96.7\%}  & {\bf 68.0\%} \\
  \hline
\end{tabular}
\caption{Performance of TransE, KRE, IIKE, TME and JRME on the metrics of Average Rank, Hit@10 and Hit@1 in NELL-5M dataset.}
\end{table}

\section{Conclusion}
We engage in bridging the gap between unstructured free texts and structured knowledge bases to predict more accurate relations via proposing a joint embedding model between any given entity pair for knowledge population. The results of extensive experiments with various evaluation protocols on both medium and large NELL datasets effectively demonstrate that our model (JRME) outperforms other state-of-the-art approaches. Because of the uniform low-dimensional vector representations for entities, relations and even the words, evidence for prediction is compressed into embeddings to facilitate the information exchange and computing, which finally leads a huge leap forward in relation extraction.

There still remain, however, several open questions on this promising research direction in the future, such as exploring better ways to embed the whole beliefs or mentions without losing too much regularities of knowledge and linguistics.
\section*{Acknowledgments}
The first author conducted this research while he was a joint-supervision Ph.D. student in New York University. This paper is dedicated to all the members of the Proteus Project.
\bibliographystyle{acl}
\bibliography{references}
\end{document}